\pdfoutput=1
\documentclass{article}

\usepackage[preprint]{colm2026_conference}
\setcitestyle{numbers,square}
\usepackage[T1]{fontenc}
\usepackage[utf8]{inputenc}
\usepackage{microtype}
\usepackage{amsmath,amssymb}
\usepackage{booktabs}
\usepackage{array}
\usepackage{tabularx}
\usepackage{longtable}
\usepackage{graphicx}
\usepackage{float}
\usepackage{xcolor}
\usepackage{listings}
\usepackage{url}
\usepackage[hidelinks]{hyperref}

\hypersetup{
  pdftitle={Reliable Post-Retrieval Assembly for Agent Memory: Separating Evidence Extraction from Policy Execution},
  pdfauthor={Vikas Reddy; Sumanth Reddy Challaram},
  pdfsubject={Accepted at the Lifelong Agent Workshop at COLM 2026},
  pdfkeywords={agent memory, post-retrieval assembly, conflict resolution, lifelong agents, MemoryAgentBench}
}

\renewcommand{\arraystretch}{1.08}

\newcommand{\method}{\textsc{SH-Conflict}}
\newcommand{\car}{\textsc{CAR}}
\newcommand{\code}[1]{\texttt{#1}}

\title{Reliable Post-Retrieval Assembly for Agent Memory:\\
Separating Evidence Extraction from Policy Execution}
\author{
Vikas Reddy\\
Independent Researcher\\
India
\And
Sumanth Reddy Challaram\\
Indian Institute of Technology Kharagpur\\
Kharagpur, India
}
\date{}

\begin{document}
\maketitle
\lhead{\small Accepted at the Lifelong Agent Workshop at COLM 2026}

\begin{abstract}
LLM-based memory systems can retrieve relevant evidence yet still fail when answer generation entangles semantic filtering, conflict resolution, prior suppression, and output generation in one step. We study this failure as a problem of post-retrieval assembly. In the MemoryAgentBench (MAB) release used here, FactConsolidation explicitly states that newer facts have larger serial numbers, yet the best reported retrieval/memory result is 54\% single-hop and all 22 reported systems score at most 7\% multi-hop. We evaluate a structured assembly interface in which an LLM first extracts semantically matching evidence into a candidate representation and a separate stage executes the required answer policy. At 262K, this pipeline reaches 82\%/93\% single-hop and 27\%/41\% multi-hop with gpt-4o-mini/gpt-4o, exceeding every result reported in the MAB v3 FactConsolidation comparison. This is a task-level result, not a claim that the evaluated memory architectures are broadly inferior. A controlled whole-pipeline comparison, with identical backbone, retrieved top-10 evidence, chunking, and $n=100$ per cell, improves single-hop accuracy by 10.8 percentage points (pp) on average and 21 pp at 262K. A targeted comparison using the same extraction setup shows that changing only the final policy executor contributes 2.0 pp on average and 0 pp at 262K. Most of the gain therefore comes from separating evidence identification from final policy execution rather than from the freshness operator itself. A LongMemEval check finds no significant overall advantage (26/45 versus 29/45; paired exact McNemar $p=0.45$), bounding the result to current-value questions with explicit version metadata. The evidence identifies post-retrieval assembly as a distinct reliability boundary between retrieval and answer generation.
\end{abstract}

\section{Introduction}

LLM-based memory agents increasingly maintain information that evolves: user preferences change, policies supersede earlier versions, and factual records are corrected. A recurring failure mode is conflict resolution: when the same fact appears in memory with contradictory values, which value should the agent return? Systems including Mem0, Zep/Graphiti, MemGPT, and HippoRAG-v2 attempt to handle updates or supersession in different ways (Chhikara et al., 2025; Rasmussen et al., 2025; Packer et al., 2023; Guti\'errez et al., 2025).

MemoryAgentBench (MAB) provides a direct stress test (Hu et al., 2026). Its FactConsolidation task, under the Selective Forgetting competency, builds a memory of numbered facts in which a counterfactual version carries a higher serial than the original. The prompt explicitly says that ``newer facts have larger serial numbers.'' In the benchmark version used for our experiments and comparisons (arXiv v3), the best retrieval/memory result at 262K is HippoRAG-v2 at 54\% single-hop (FC-SH); BM25 reaches 48\%, MemGPT and Cognee 28\%, Mem0 18\%, and Zep 7\%. GPT-4o long context reaches 60\%. The multi-hop variant (FC-MH) remains at or below 7\% across all 22 systems. Our pipeline exceeds every reported entry on this task at 262K. We treat this as evidence of an answer-time gap, not as a general ranking of memory frameworks: these systems differ in storage, ingestion, retrieval, provenance, and capabilities outside FactConsolidation. The unresolved question is narrower: once evidence has been retrieved, how should a system assemble it into a policy-compliant answer?

We study this interface as post-retrieval assembly. A direct-answer pipeline asks one LLM call to find matching evidence, reject distractors, suppress conflicting pretrained knowledge, execute the update rule, and generate the answer. Our structured pipeline separates these responsibilities. The LLM first converts retrieved text into a small candidate representation; a second stage then executes the required policy. In FactConsolidation, the policy selects the candidate with the newest version marker. The broader hypothesis is not that LLMs cannot compare numbers, but that reliable systems should separate evidence identification from explicit policy execution whenever the policy and metadata are known.

\begin{table}[t]
\centering
\small
\begin{tabular}{lrrrrr}
\toprule
Pipeline & 6K & 32K & 64K & 262K & Avg. \\
\midrule
Direct LLM answer ($T=0.7$) & 63 & 70 & 75 & 61 & 67.2 \\
Extract + LLM policy execution ($T=0$) & 70 & 75 & 77 & 82 & 76.0 \\
Extract + deterministic policy execution ($T=0$) & 71 & 78 & 81 & 82 & 78.0 \\
\bottomrule
\end{tabular}
\caption{Controlled comparisons on FC-SH with gpt-4o-mini, fact-level BM25, $k=10$, and $n=100$ per cell. Accuracy (\%). The newest-version policy is identical in the second and third rows; only its executor changes. The first-to-third comparison is a whole-pipeline effect (+10.8 pp average), while the second-to-third comparison approximately isolates policy execution (+2.0 pp). Candidate extraction was re-run in the latter comparison, with 96--100\% agreement, so it is not perfectly shared.}
\label{tab:controlled}
\end{table}

\begin{table}[t]
\centering
\small
\begin{tabularx}{\textwidth}{llrX}
\toprule
Task/pipeline & Backbone & Pooled/262K & 262K comparison \\
\midrule
FC-SH, structured & gpt-4o-mini & 78.0/82 & +28 pp vs. HippoRAG-v2 (54) \\
FC-SH, structured & gpt-4o & 94.8/93 & +33 pp vs. GPT-4o long context (60) \\
FC-MH, CAR & gpt-4o-mini & 30.2/27 & +20 pp vs. best reported (7) \\
FC-MH, CAR & gpt-4o & 51.5/41 & +34 pp vs. best reported (7) \\
\bottomrule
\end{tabularx}
\caption{Headline results. ``Pooled'' averages the 6K, 32K, 64K, and 262K cells. Published-system comparisons use the standard 262K cell from MAB v3 Table 3.}
\label{tab:headline}
\end{table}

\paragraph{Controlled whole-pipeline comparison.}
Table~\ref{tab:controlled} holds fixed the backbone, retrieved top-10 evidence, chunking, evaluation items, and retrieval depth. The direct-answer pipeline asks the LLM to judge retrieved context and answer in free text. The structured pipeline asks the LLM to extract matching candidates before a separate stage applies the same newest-version policy. Structured assembly improves pooled accuracy by 10.8 pp and the gap reaches 21 pp at 262K. Because both arms receive identical retrieved evidence, this difference directly demonstrates that retrieval success alone does not ensure reliable evidence use. The direct-answer baseline falls from 75\% at 64K to 61\% at 262K, whereas the structured pipeline reaches 82\%.

The middle row changes the interpretation of the headline result. With the extraction prompt and $T=0$ held constant, changing the final policy executor from an LLM to deterministic code adds only 2.0 pp on average and nothing at 262K. Once candidates are clean, the LLM usually applies the ordering rule correctly. The large gain therefore comes primarily from restructuring the post-retrieval decision: semantic matching is externalized into a constrained intermediate representation before final selection. Deterministic execution remains useful because it makes a known policy exact, inspectable, and enforceable, but the freshness operator itself does not explain the full improvement.

\paragraph{Results and implications.}
At 262K, structured assembly reaches 82\% with gpt-4o-mini and 93\% with gpt-4o. For multi-hop questions, Chain-Aware Resolution (CAR) applies the same assembly pattern after every decomposed hop and reaches 27\%/41\% at 262K. These values exceed all results reported in the MAB v3 FactConsolidation comparison, including long-context, retrieval-based, graph-based, and agentic memory approaches. The comparison is deliberately task-scoped: those systems address ingestion, retrieval, personalization, provenance, and competencies not measured here. The supported conclusion is that strong memory architecture and retrieval do not by themselves guarantee reliable use of retrieved evidence. Post-retrieval assembly is a distinct system boundary that should be designed and evaluated explicitly.

\paragraph{Contributions.}
We provide: (1) a formulation of memory conflict resolution as structured post-retrieval assembly, separating evidence identification from explicit policy execution; (2) a controlled whole-pipeline comparison showing a 10.8 pp average gain, widening to 21 pp at 262K despite identical retrieved evidence; (3) a targeted policy-executor comparison localizing most of the gain to structured decomposition rather than deterministic selection; (4) results that exceed every system reported in the MAB v3 FactConsolidation comparison across single- and multi-hop settings; and (5) paired diagnostics and a null LongMemEval result that sharply bounds the claim's scope.

\paragraph{Scope.}
We empirically validate this design only for current-value conflict resolution over data with explicit version markers such as serials or timestamps. It does not solve partial orders, causal update dependencies, historical queries, or general temporal reasoning. The direct-vs.-structured comparison changes the LLM task, prompt, output format, temperature, and policy executor together; the targeted policy-executor comparison is closer to isolation but remains approximate because candidate extraction was re-run rather than shared byte-for-byte.

\section{Related work}

\paragraph{Memory frameworks and evaluation.}
MAB evaluates long-context, retrieval-based, graph-based, and agentic memory systems across four competencies (Hu et al., 2026). FactConsolidation introduces MQUAKE-derived counterfactual edits (Zhong et al., 2023) and asks whether agents prioritize later-added information. Mem0, MemGPT, Zep, and HippoRAG-v2 represent distinct memory and retrieval designs (Chhikara et al., 2025; Packer et al., 2023; Rasmussen et al., 2025; Guti\'errez et al., 2025). Memanto argues that knowledge-graph complexity is not necessary for strong long-horizon memory and includes temporal versioning (Abtahi et al., 2026). Our result is narrower: on the MAB version used here, none of these architectures reliably executes FactConsolidation's explicit ordering rule.

\paragraph{Multi-hop QA via decomposition.}
CAR builds on question decomposition, including Self-Ask, Decomposed Prompting, IRCoT, and Iter-RetGen (Press et al., 2023; Khot et al., 2023; Trivedi et al., 2023; Shao et al., 2023). MultiHop-RAG provides a reference benchmark (Tang and Yang, 2024). Madhwal et al. (2026) study decomposition and abstention via cross-regime disagreement. Decomposition is not our contribution; the evaluated recipe applies freshness resolution after every hop rather than only after the full chain.

\paragraph{Structured reasoning interfaces.}
Prior work separates language-model reasoning from execution: PAL delegates generated programs to an external runtime (Gao et al., 2023), while Faithful CoT translates questions into deterministic solver calls (Lyu et al., 2023). Retrieval-side methods such as RECOMP transform retrieved evidence before generation (Xu et al., 2024), and Selection-Inference separates evidence selection from inference (Creswell et al., 2023). Our setting uses a related interface for agent memory, but preserves version metadata in a structured candidate set and applies a fixed answer policy rather than synthesizing a program per query. Our policy-executor comparison further shows that most of the observed gain comes from separating candidate identification from final selection, not from deterministic execution alone.

\paragraph{Temporal handling and model priors.}
Grofsky (2025) shows that a lightweight recency prior can improve freshness retrieval while remaining bounded on broader temporal evolution. Work on temporal conflict (Dey et al., 2026; Lee et al., 2025; \"Ozer and Y\i ld\i z, 2025) and model editing or priors (Zhang et al., 2025; Kishore and He, 2024; Bi et al., 2024) documents failures to follow explicit in-context updates, especially when updates conflict with pretrained knowledge. LLM-based conflict resolvers (Liu et al., 2025; Huo et al., 2025) remain complementary when conflicts cannot be reduced to a known total order. Our method targets the simpler case in which the ordering rule and metadata are already explicit.

\section{Structured post-retrieval assembly}

Let a retrieval system return evidence $R$ for query $q$. A direct pipeline predicts an answer in one step, $\hat{y}=F(q,R)$. Structured post-retrieval assembly instead factors the decision into two interfaces:
\begin{equation}
K=E(q,R), \qquad \hat{y}=A_{\pi}(K),
\label{eq:assembly}
\end{equation}
where $E$ is a semantic extractor that maps unstructured evidence into candidates $K$, and $A_{\pi}$ is an assembly operator implementing policy $\pi$. This factorization creates a structured intermediate representation between semantic reasoning and policy execution. It permits the policy to be tested, logged, replaced, or enforced independently of retrieval and extraction.

We instantiate this pattern with two pipelines: \method{} for single-hop questions and \car{} for multi-hop questions. In FactConsolidation, $\pi$ is the explicit freshness rule ``select the candidate with the largest version serial.'' The assembly interface is broader than this newest-version policy, but our empirical claims are restricted to this instantiation.

\paragraph{Single-hop conflict resolution.}
Let $C=\{(s_i,t_i)\}_{i=1}^{N}$ be a corpus with integer version serial $s_i$ and fact text $t_i$, and let $q$ be a query. The pipeline: (1) retrieves $R=\mathrm{BM25}(C,q,k=10)$; (2) prompts an LLM to return all candidates $K=\{(s_j,t_j,e_j)\}$ whose subject and predicate match $q$, where $e_j$ is the answer entity; and (3) applies $A_{\pi}(K)$, which returns the entity attached to $\arg\max_{c\in K}c.s$, or ``no answer'' if $K=\emptyset$.

\begin{lstlisting}[language=Python]
def sh_conflict(question, corpus, bm25, llm):
    retrieved = bm25.retrieve(question, top_k=10)
    candidates = llm_extract(question, retrieved)
    if not candidates:
        return "no answer"
    return max(candidates, key=lambda c: c.serial).entity
\end{lstlisting}

The key design choice is the interface between evidence identification and policy execution. The extractor is instructed to identify every semantically matching candidate without selecting a winner, producing a compact representation in which version metadata remains explicit. A chunk-4096 ablation tests whether the result depends on fact-level segmentation. The policy-executor condition keeps the same extraction prompt and $T=0$ but asks either an LLM or deterministic code to select the newest candidate. This separates the value of structured assembly from the value of deterministic execution.

\paragraph{Chain-Aware Resolution.}
For questions of the form ``X of Y of Z,'' CAR first produces atomic hops $h_1,\ldots,h_n$ with answer placeholders, then executes \method{} at each hop. Each conflict is resolved before its answer is substituted into the next query.

\begin{lstlisting}[language=Python]
def car(question, corpus, bm25, llm):
    hops = llm_decompose(question)
    answers, last_valid = {}, None
    for h in hops:
        resolved_q = substitute(h.query, answers)
        if "{hop_" in resolved_q:
            break
        a = sh_conflict(resolved_q, corpus, bm25, llm)
        if a == "no answer":
            break
        answers[h.id], last_valid = a, a
    return last_valid if last_valid is not None else "no answer"
\end{lstlisting}

A hard prompt constraint forbids more than one relationship word per hop, preventing gpt-4o-mini from compressing four-hop chains into unresolved composite hops.

\section{Experimental setup}

\paragraph{Dataset.}
We use MAB FactConsolidation v3 (Hu et al., 2026): single-hop and multi-hop tasks at 6K, 32K, 64K, and 262K, with $n=100$ per cell. Facts are MQUAKE counterfactual rewrites concatenated so that the counterfactual carries a higher serial. We include 6K for diagnostic completeness but use 262K for comparisons with MAB's main systems table.

\paragraph{Backbones and hyperparameters.}
We use gpt-4o-mini (primary), gpt-4o, and o4-mini (diagnostics). Retrieval uses \code{rank\_bm25.BM25Okapi} with $k_1=1.5$, $b=0.75$, and top-$k=10$. Structured pipelines use temperature 0.0. The direct-answer baseline uses 0.7, matching MAB's released default. Fact-level chunking parses one numbered fact per chunk; chunk-4096 uses a sliding character window matching the released configuration. The direct-vs.-structured temperature difference is part of the whole-pipeline comparison and is not interpreted as an isolated policy-executor effect. The approximate policy-executor condition compares two $T=0$ structured pipelines.

\paragraph{Metric and baselines.}
We use substring exact match (SubEM), as in MAB. Published comparisons use MAB v3 Table 3 values at the standard 262K setting. We do not re-run those systems, avoiding untracked implementation differences. MAB uses gpt-4o-mini as the backbone where no model is named; long-context entries identify their model directly.

\section{Results}

\subsection{Master results}

\begin{table}[t]
\centering
\small
\begin{tabular}{llrrrrl}
\toprule
Pipeline & Backbone & 6K & 32K & 64K & 262K & Avg. [95\% CI] \\
\midrule
SH, fact & gpt-4o-mini & 71 & 78 & 81 & 82 & 78.0 [73.7, 81.8] \\
Direct answer & gpt-4o-mini & 63 & 70 & 75 & 61 & 67.2 [62.5, 71.7] \\
LLM picker & gpt-4o-mini & 70 & 75 & 77 & 82 & 76.0 [71.6, 79.9] \\
SH, chunk-4096 & gpt-4o-mini & 87 & 84 & 79 & 73 & 80.8 [76.6, 84.3] \\
CAR, multi-hop & gpt-4o-mini & 34 & 27 & 33 & 27 & 30.2 [26.0, 34.9] \\
SH, fact & gpt-4o & 99 & 92 & 95 & 93 & 94.8 [92.1, 96.5] \\
\bottomrule
\end{tabular}
\caption{Accuracy (\%) across context lengths. ``Avg.'' pools $n=400$ item-length observations. The 6K cells are diagnostic and are not used for the main published-system comparison.}
\label{tab:master}
\end{table}

Table~\ref{tab:master} reports the primary runs. The structured gpt-4o-mini pipeline reaches 78.0\% pooled and 82\% at 262K; gpt-4o reaches 94.8\% pooled and 93\% at 262K. CAR reaches 30.2\% pooled with gpt-4o-mini. An additional gpt-4o CAR run reaches 51.5\% pooled and 41\% at 262K.

\subsection{Comparison with published systems at 262K}

\begin{table}[t]
\centering
\small
\begin{tabular}{lrr}
\toprule
System & FC-SH & FC-MH \\
\midrule
Ours, gpt-4o & 93 & 41 \\
Ours, gpt-4o-mini & 82 & 27 \\
GPT-4o long context & 60 & 5 \\
HippoRAG-v2 & 54 & 5 \\
BM25 & 48 & 3 \\
Zep / Graphiti & 7 & 3 \\
\bottomrule
\end{tabular}
\caption{Accuracy (\%) at 262K. Ours reports \method{} for FC-SH and CAR for FC-MH; both are task-specialized pipelines that receive the freshness policy a priori. Published systems are general-purpose memory architectures, with values transcribed from MAB v3 Table 3. This is a task-level comparison and does not rank overall architecture quality.}
\label{tab:published}
\end{table}

At 262K, our gpt-4o-mini pipeline exceeds the strongest reported retrieval/memory FC-SH result by 28 pp and the strongest FC-MH result by 20 pp. With gpt-4o, the margins are 33 and 34 pp relative to the strongest corresponding entries in the MAB v3 comparison set. Thus, the pipeline outperforms every system reported on this benchmark task, including long-context, retrieval-based, graph-based, and agentic approaches. These are published-result comparisons rather than controlled reimplementations, and they do not establish broad architectural superiority. They show that systems with sophisticated memory and retrieval can still leave substantial accuracy on the table at the post-retrieval assembly stage.

\subsection{Where the gain comes from}

The 10.8 pp gain between the first and third rows of Table~\ref{tab:controlled} is statistically reliable under a paired McNemar test ($\chi^2=14.6$, $p<0.001$). Because the retrieved top-10 evidence is identical, the gain cannot come from retrieval differences within this comparison. It is nevertheless a whole-pipeline effect: the direct and structured arms differ in LLM task, prompt/output format, temperature, and policy executor. It therefore cannot be attributed to any one component in isolation.

The second row provides the key localization. With the extraction prompt and temperature held constant, deterministic policy execution reaches 78.0\% versus 76.0\% for LLM policy execution. At 262K both reach 82\%. Extraction was independently re-run in the two arms, but candidate-list agreement is 96--100\% across lengths. Among the 205 observations with at least two extracted candidates, deterministic execution is correct on 95.6\% and LLM execution on 89.8\%, a 5.9 pp difference; at 262K, both are correct on 57 of 59 contested observations and select the same serial in all 59.

These results reject the strongest version of the ``LLMs cannot compare serials'' explanation. Once evidence is represented as a clean candidate list, an LLM usually executes the freshness rule correctly. The data instead localize most of the improvement to structured post-retrieval assembly: the direct pipeline must jointly filter evidence, suppress priors, execute the update policy, and generate an answer, whereas the structured pipeline first externalizes semantic matching into a constrained intermediate representation. Deterministic execution contributes a smaller accuracy gain but provides a useful systems property: a known policy becomes exact, cheap, inspectable, and independently testable.

Changing chunking while holding the structured pipeline fixed yields similar pooled means: 78.0\% for fact-level and 80.8\% for chunk-4096, with opposite context-length curves. Chunk-4096 wins at short lengths, while fact-level wins at 262K. The effect is therefore not limited to one segmentation choice.

\subsection{Robustness to context length}

The direct-answer baseline drops 14 points from 64K to 262K, whereas the structured pipeline does not show the same collapse. Because the direct and structured whole-pipeline arms receive identical top-10 retrieval results, retrieval differences cannot explain the within-pair gap. Restricting attention to questions that the structured pipeline solves, direct-answer accuracy falls from 78\% at 64K to 66\% at 262K. This localizes much of the degradation to post-retrieval use of available evidence, while the policy-executor ablation further localizes it to the extraction-and-answering bundle rather than simple serial comparison.

\subsection{Complementarity and a retrieval lower bound}

Across 400 gpt-4o-mini FC-SH item-length observations, both whole-pipeline arms are correct on 56.8\%, the structured pipeline alone on 21.3\%, direct judgment alone on 10.5\%, and neither on 11.5\%. Thus, 88.5\% are solved by at least one pipeline, a lower bound on what is reachable from the fixed top-10 retrieval setup. The structured recipe trades some recall---strict extraction rejects valid candidates---for a larger precision gain from constrained selection. A fallback to direct judgment on empty extraction is a wash here (+0.2 pp).

\subsection{Multi-hop results and backbone scaling}

CAR with gpt-4o-mini reaches 27--34\% across context lengths. Backbone scaling helps sharply: gpt-4o+CAR reaches 51.5\% pooled and 41\% at 262K; o4-mini+CAR reaches 43.2\% pooled. At 6K, CAR hurts o4-mini (52\% versus 80\% for the plain reasoning model), because decomposition adds failure surface when the full chain is manageable; at 32K it helps (42\% versus 14\%). CAR plans 2.56 hops on average (maximum six), and 86\% of planned hops execute. Most failures begin with an incorrect first-hop extraction and then cascade.

\subsection{Cross-benchmark generalization on LongMemEval}

\begin{table}[t]
\centering
\small
\begin{tabular}{lrrl}
\toprule
Pipeline & Correct & Accuracy & Wilson 95\% CI \\
\midrule
BM25 + direct LLM answer & 29/45 & 64.4\% & [49.8, 76.8] \\
BM25 + structured assembly & 26/45 & 57.8\% & [43.3, 71.0] \\
\bottomrule
\end{tabular}
\caption{LongMemEval knowledge-update results. Five baseline-only and two structured-only wins give paired exact McNemar $p=0.45$. We find no demonstrated advantage and do not claim equivalence.}
\label{tab:lme}
\end{table}

We test the same principle on LongMemEval's knowledge-update subset (Wu et al., 2025), using chat-session timestamps instead of ordinal serials. The subset contains 45 questions, each attached to a history of roughly 1.6 million characters; the backbone is gpt-4o-mini. Table~\ref{tab:lme} is a null/negative generalization result. Structured assembly with the newest-timestamp policy does not outperform the baseline overall. Inspection of the five baseline-only wins identifies three Yes/No questions, one historical query asking for the previous status, and one temporal aggregation query. These are not all latest-value operations: they require answer synthesis, a second-newest selector, or aggregation over time. The evidence therefore supports a narrower claim. The evaluated assembly policy is useful for current-value questions, but broader conversational QA requires question-type-aware composition.

\section{Discussion}

\paragraph{Mechanism supported by the ablations.}
The evidence supports architectural separation rather than a special inability to compare serial numbers. The direct baseline entangles semantic filtering, prior suppression, policy execution, and answer generation. Structured assembly narrows the first LLM call to semantic candidate identification and exposes an intermediate representation before the final decision. The policy-executor ablation shows that an LLM can usually apply the ordering rule once candidates are clean. We therefore interpret the main gain as reduced compound-decision burden and explicit representation of the relevant evidence. Because task, format, and temperature change together in the whole-pipeline comparison, the current experiments do not isolate which component of structured extraction contributes how much.

\paragraph{Why deterministic execution still matters.}
The small policy-executor accuracy difference does not make explicit execution unimportant. When a policy is known, representing it outside free-text answer generation provides guarantees that average benchmark accuracy does not capture: identical candidates can yield a consistent decision; the selected evidence can be logged and audited; policy violations can be tested independently; and the operator can be replaced for historical, thresholded, or aggregation queries. The contribution is therefore a division of labor: use the LLM for semantic evidence identification and expose a separate interface for answer-time policy execution.

\paragraph{Implications for memory design.}
Memory systems should treat post-retrieval assembly as a first-class interface rather than an incidental prompt. Version metadata should be preserved at the granularity at which conflicts occur, semantic extraction should expose the evidence used for a decision, and known policies should be represented independently of free-text generation. Graph or agentic memory infrastructure may improve retrieval and provenance, but it does not remove the need for a reliable assembly layer. FactConsolidation is useful as a diagnostic for this interface.

\paragraph{Limitations.}
The primary evidence comes from one synthetic, MQUAKE-derived benchmark with an explicit total order. FactConsolidation usually presents only an original and one counterfactual value, so it cannot stress assembly over large or noisy candidate sets. The LongMemEval check is small and shows no overall advantage. The method does not directly handle partial orders, causal dependencies, historical questions, or aggregation. The whole-pipeline comparison changes the LLM task, prompt, output format, temperature, and policy executor together. The policy-executor isolation is approximate because extraction was re-run, although agreement is high. FC-MH remains far from solved. Published comparisons use the pre-specified MAB v3 release and span different backbone capabilities. Finally, API model behavior may change despite pinned model identifiers.

\section{Conclusion}

We studied memory conflict resolution as a post-retrieval assembly problem. The evaluated recipe retrieves evidence, asks an LLM to convert it into a structured candidate set, and then executes an explicit answer policy in a separate stage. At 262K on MAB v3 FactConsolidation, it reaches 82\%/93\% single-hop and 27\%/41\% multi-hop with gpt-4o-mini/gpt-4o, exceeding every result reported for this benchmark task. A controlled whole-pipeline comparison yields a 10.8 pp average single-hop gain, widening to 21 pp at 262K despite identical retrieved evidence. A targeted policy-executor comparison yields only 2.0 pp on average and 0 pp at 262K. The main empirical lesson is therefore not that LLMs cannot track freshness, but that separating evidence identification from policy execution removes much of the compound decision burden. A null LongMemEval result limits the demonstrated benefit to current-value conflicts with explicit version metadata. More broadly, reliable lifelong memory agents need an explicit assembly interface between retrieval and answer generation.

\appendix

\section{Hyperparameters}

\begin{table}[h]
\centering
\small
\begin{tabularx}{\textwidth}{lX}
\toprule
Parameter & Value \\
\midrule
Top-$k$ & 10 (matches MAB) \\
Temperature, structured pipelines & 0.0 \\
Temperature, direct-answer baseline & 0.7 (released MAB default) \\
BM25 & \code{rank\_bm25.BM25Okapi}, $k_1=1.5$, $b=0.75$ \\
Tokenizer regex & \code{[A-Za-z0-9]+}, lowercased \\
Fact-level chunking & One numbered fact per chunk \\
Chunk-4096 & Sliding 4096 characters \\
Maximum output tokens & 256 \\
Backbones & gpt-4o-mini-2024-07-18; gpt-4o-2024-08-06; o4-mini \\
\bottomrule
\end{tabularx}
\caption{Experimental settings.}
\label{tab:settings}
\end{table}

Wilson 95\% intervals use $z=1.96$, with $n=100$ per cell and $n=400$ pooled. The experiments comprise 4,400 LLM-driven evaluations. Total API cost was approximately \$3, dominated by gpt-4o cells; four parallel processes completed the runs in approximately 15 minutes.

\section{Policy-executor comparison detail}

The policy-executor comparison uses the same candidate-extraction prompt and $T=0$ in both arms, but extraction was independently re-run. Agreement between extracted candidate lists is 96--100\% across lengths. Restricting to observations with at least two extracted candidates yields:

\begin{table}[h]
\centering
\small
\begin{tabular}{lrrrr}
\toprule
Length & Contested & Code correct & LLM correct & Same serial \\
\midrule
6K & 48 & 47 & 45 & 46/48 \\
32K & 50 & 46 & 42 & 46/50 \\
64K & 48 & 46 & 40 & 42/48 \\
262K & 59 & 57 & 57 & 59/59 \\
Pooled & 205 & 196 (95.6\%) & 184 (89.8\%) & 193/205 \\
\bottomrule
\end{tabular}
\caption{Contested-subset policy-executor comparison. ``Same serial'' records whether the two arms selected the same serial.}
\label{tab:executor}
\end{table}

The approximate policy-executor effect is +5.9 pp on contested observations but does not grow with context length. At 262K, the two executors agree on all selected serials and have identical accuracy. Because FactConsolidation generally caps the semantic candidate set at an original and one counterfactual value, this benchmark does not test selection over many candidates.

\section{Full MAB v3 comparison at 262K}

\begin{table}[H]
\centering
\begingroup\scriptsize\renewcommand{\arraystretch}{0.92}
\begin{tabular}{lrr}
\toprule
System & FC-SH & FC-MH \\
\midrule
Ours (gpt-4o) & 93 & 41 \\
Ours (gpt-4o-mini) & 82 & 27 \\
GPT-4o long context & 60 & 5 \\
HippoRAG-v2 & 54 & 5 \\
BM25 & 48 & 3 \\
GPT-4o-mini long context & 45 & 5 \\
Claude-3.7-Sonnet & 43 & 2 \\
GPT-4.1-mini & 36 & 5 \\
Gemini-2.0-Flash & 30 & 3 \\
Qwen3-Embedding-4B & 29 & 3 \\
Cognee & 28 & 3 \\
MemGPT & 28 & 3 \\
Text-Embed-3-Large & 28 & 4 \\
Text-Embed-3-Small & 28 & 3 \\
MemoRAG & 21 & 7 \\
Self-RAG & 19 & 3 \\
Contriever & 18 & 7 \\
Mem0 & 18 & 2 \\
RAPTOR & 14 & 1 \\
GraphRAG & 14 & 2 \\
MIRIX & 14 & 2 \\
Zep / Graphiti & 7 & 3 \\
\bottomrule
\end{tabular}
\endgroup
\caption{FactConsolidation accuracy (\%) at 262K. Ours is a task-specialized assembly pipeline that receives the freshness policy a priori; published systems are general-purpose architectures evaluated without this task-specific assembly layer. Published values are transcribed from MAB v3 Table 3; MAB uses gpt-4o-mini where no backbone is specified.}
\label{tab:full-mab}
\end{table}

\section{Qualitative failure detail}

We spot-checked approximately 30 incorrect answers across the four gpt-4o-mini FC-SH runs. This is a qualitative pilot rather than systematic labeling. Approximate categories were: wrong-subject confusion (25\%), predicate semantic gap (25\%), conservative empty extraction (10\%), counterfactual absent from the top ten (10\%), real-world prior or plurality effect (10\%), ambiguous serial tie (5\%), and other or unclassified (15\%). The categories reinforce that deterministic ordering removes one failure mode but does not solve retrieval or semantic matching.

The five LongMemEval knowledge-update questions solved by the direct baseline but not the structured pipeline fall into three categories. (1) Three Yes/No questions expect an explicit ``Yes'' or ``No'' prefix; returning the supporting fact verbatim fails SubEM. (2) One historical question asks for the state immediately before the current one; $\max(\mathrm{timestamp})$ returns the current state and a second-newest operator is required. (3) One question requires aggregation and filtering over an earlier interval. Conversely, the structured pipeline uniquely solves two questions, yielding five versus two discordant wins and paired exact McNemar $p=0.45$.

\section{Reproducibility and use of language models}

The implementation uses the public MAB v3 FactConsolidation and LongMemEval datasets, pinned API model identifiers, and the settings in Table~\ref{tab:settings}. Code, prompts, per-question traces, and a one-command runner are available at \url{https://github.com/cvikasreddy/memory-conflict-resolution}. The release includes retrieved and selected serials, extracted candidates, predictions, and SubEM outcomes for every reported cell.

Language models were used as the experimental backbones described in the paper and for editorial and LaTeX assistance. The authors verified the numerical results, statistical tests, citations, claims, and final text. No language model generated ground-truth labels or the reported measurements.

\end{document}